\documentclass{cta-author}

\usepackage{booktabs} 
\usepackage{graphicx}
\usepackage{epstopdf}
\usepackage{subfigure}
\usepackage{amsmath,amssymb,amsfonts}
\usepackage{xcolor}
\usepackage{enumerate}
\usepackage{amsthm}
\usepackage{diagbox}
\usepackage{enumitem} 
\usepackage{color,colordvi}
\usepackage{textcomp}
\usepackage{epsfig}
\usepackage{url}
\usepackage{mdwlist}
\usepackage{makecell}
\usepackage{verbatim}
\usepackage{multirow,multicol}
\usepackage{setspace}
\usepackage{float}
\usepackage{fancyhdr}
\usepackage[normalem]{ulem}
\usepackage{makecell}
\usepackage{fancyhdr}
\usepackage{flushend}
\usepackage{algorithm}
\usepackage{algorithmic}
\usepackage{natbib}
\usepackage{multirow}
\usepackage{soul}
\usepackage[colorlinks,linkcolor=green,citecolor=green]{hyperref}
{}
{}
{}

\begin{document}
\title{Self-supervised Multi-view Clustering in Computer Vision: A Survey}

\author{
\au{Jiatai Wang$^{1,2}$}, \au{Zhiwei Xu$^{3,4\corr}$}, \au{Xuewen Yang$^{5}$},  \au{Hailong Li$^{1}$}, \au{Bo Li$^{1}$}, \au{Xuying Meng$^{4}$}
}

\address{
\add{1}{College of Data Science and Application, Inner Mongolia University of Technology, Huhhot, China, 010080}
\add{2}{Voice Semantics Research Department, OPPO Research Institute, Beijing, China, 100020}
\add{3}{Haihe Laboratory of Information Technology Application Innovation, Tianjin, China,  300350}
\add{4}{Institute of Computing Technology, Chinese Academy of Sciences, Beijing, China, 100190}
\add{5}{InnoPeak Technology, Inc, CA, USA, 94303}
\email{xuzhiwei2001@ict.ac.cn}
}

\begin{abstract}
Multi-view clustering (MVC) has had significant implications in cross-modal representation learning and data-driven decision-making in recent years. It accomplishes this by leveraging the consistency and complementary information among multiple views to cluster samples into distinct groups. However, as contrastive learning continues to evolve within the field of computer vision, self-supervised learning has also made substantial research progress and is progressively becoming dominant in MVC methods. It guides the clustering process by designing proxy tasks to mine the representation of image and video data itself as supervisory information. Despite the rapid development of self-supervised MVC, there has yet to be a comprehensive survey to analyze and summarize the current state of research progress. Therefore, this paper explores the reasons and advantages of the emergence of self-supervised MVC and discusses the internal connections and classifications of common datasets, data issues, representation learning methods, and self-supervised learning methods. This paper does not only introduce the mechanisms for each category of methods but also gives a few examples of how these techniques are used. In the end, some open problems are pointed out for further investigation and development.
\end{abstract}

\maketitle

\section{Introduction}\label{sec1}

Data often presents multiple views, collected from diverse sensors or obtained through various feature extractors. For example, specific news events are reported by multiple news organizations, RGB images or depth maps are captured by different types of cameras or from varying angles by the same camera, and videos can take on multiple forms, including images, audio, and text. Consequently, single-view methods struggle to effectively utilize the information contained within multi-view data. To better construct a comprehensive vision model of an object, it is essential to comprehensively observe its various views or utilize multiple modalities within images and videos.
Hence, there is a strong demand for effective multi-view learning methods, especially those that operate in an unsupervised manner, in real-world vision applications. 
As one of the most important unsupervised multi-view methods, multi-view clustering (MVC) aims to separate data points into different clusters in an unsupervised fashion \cite{153,154,155,156,157,158}. To achieve this end, existing methods \cite{4,75,6,95,96,146} use deep neural networks to explore consistency and complementarity across different views so that a common/shared representation is learned. However, some deep MVC methods \cite{18,19,24,65, 66} depend on too many hyperparameters.
In practical clustering applications, where label information is lacking for tuning, this poses a significant challenge. Furthermore, many deep MVC methods suffer from shortcomings such as limited representation capability and high computational complexity, thereby constraining their performance when tackling large-scale data clustering tasks.

\begin{figure}[t!]
\centering
\includegraphics[width=1\linewidth]{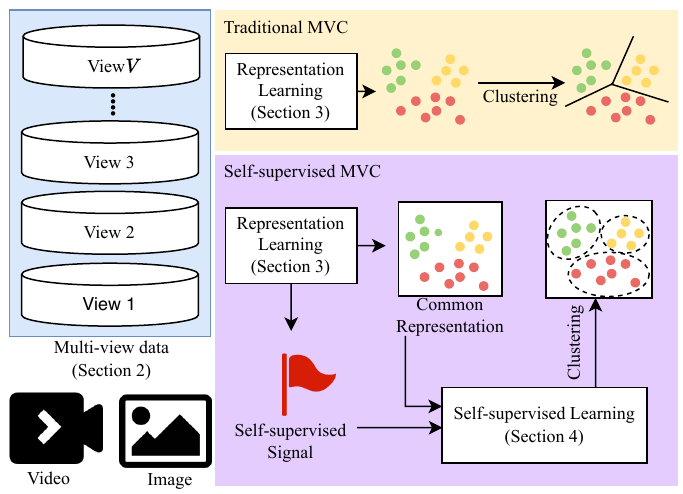}
\caption{Process illustration of self-supervised MVC and traditional MVC. Images or videos are the source of multi-view data. Based on these data, traditional MVC methods will first do representation learning to get high-dimensional semantic features and then complete clustering by using clustering algorithms such as K-means \cite{159}. Self-supervised MVC methods will generate self-supervised signals, e.g., pseudo-labels, through representation learning and then mine the consistency of views through self-supervised learning methods, such as contrastive learning, and thus have stronger generalization ability and robustness.}
\label{fig1}
\end{figure}
\begin{figure*}[h!]
  \begin{center}
  \includegraphics[width=1\textwidth]{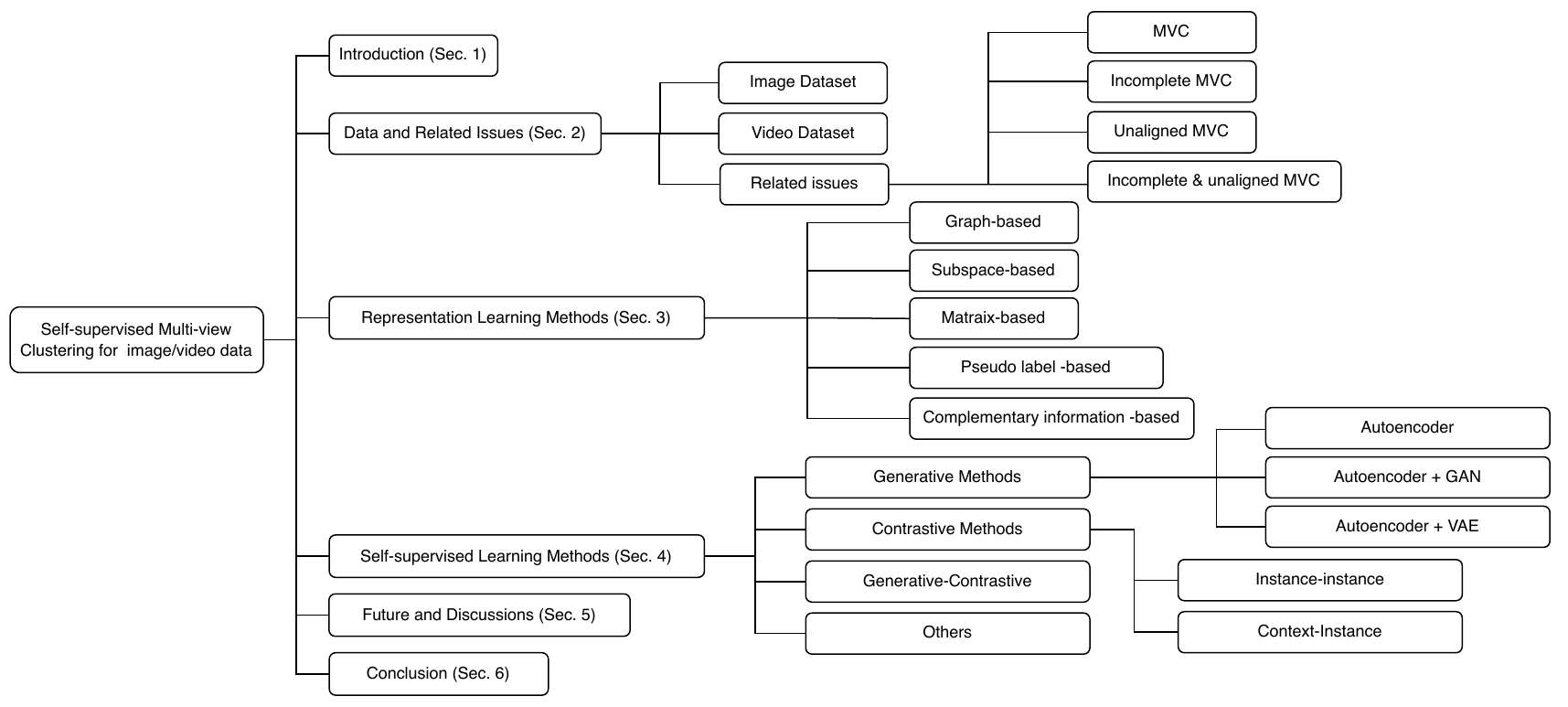}
  \caption{Structure of this paper.
  }\label{fig2}
  \end{center}
\end{figure*}
To overcome the above issues, self-supervised multi-view learning based methods are gradually appearing in some works \cite{14, 15, 16, 17,18}, and better progress has been made in guiding the feature learning process through the self-supervised signal. It guides all multiple views to learn more distinguishing features by extracting common representations from view-specific representations and keeping each view space independent, thus minimizing the distance between positive samples and maximizing the distance between negative samples, and forming pseudo-labels. The pseudo-label as a self-supervised signal can be used to lead all views to learn more discriminative features, which further produce clearer clustering structures, as shown in Fig. \ref{fig1}. Therefore, self-supervised multi-view clustering pre-trains sample data to obtain pseudo-labels and achieves good supervision and guidance for downstream clustering tasks through migration and fine-tuning \cite{20,25,74,85,86}. Additionally, the diverse learning methods and self-supervised signal representations bring both opportunities and difficulties for future research. Self-supervised MVC has thus attracted more and more attention in the past few years, which makes it necessary and beneficial to summarize the state of the art and delineate open problems to guide future advancement.

The prevailing method in the current paradigm is to perform linear clustering following representation learning. This is achieved by sampling from multi-view data distributions in an auto-encoding reconstruction process to extract self-supervised signals \cite{4, 5, 161}. Additionally, it's important to note that self-supervised signals are not limited to pseudo-labels; they extend to various abstract models or mathematical functions used in representation learning, which we will describe in detail below. In essence, self-supervised learning is rooted in transfer learning \cite{162}, implying that this method transfers features from the lower levels of multi-view data rather than semantic features from the top. Thus, we can preserve better spatial location relations by using artificially designed proxy tasks \cite{121}, leading to correct representation learning in training.
In this paper, we categorize existing work from three perspectives: data challenges, representation learning, and self-supervised learning:
a) Missing and unaligned data is the primary problem and major challenge for self-supervised MVC.
b) Representation learning determines the features of self-supervised signaling and needs to be revisited.
c) Self-supervised learning methods are the main means of consistency learning. For example, some generative methods use Generative Adversarial Network (GAN) \cite{101} to deal with missing data, and some contrastive methods are more widely used based on the idea of instance discrimination.
On one hand, previous works \cite{75,132,133} primarily focused on reviewing existing shallow and deep model-based MVC without delving into self-supervised learning. In our survey, we give self-supervised MVC a more focused exploration, considering it as the future mainstream. On the other hand, by analyzing and comparing their technical details, we discuss the current challenges and future directions. Understanding their commonalities will benefit researchers in developing a more unified and collaborative framework, ultimately methoding the complexity of real human intelligence systems.

Several comprehensive reviews have been conducted in the domains of multi-view clustering \cite{32, 8}, incomplete multi-view clustering \cite{87, 160}, and representation learning within multi-view clustering \cite{11}. However, none of these reviews have focused on the intriguing concept of self-supervised learning itself. To the best of our knowledge, there has been no comprehensive survey and summary of the field of self-supervised multi-view clustering concerning image and video data. This gap in the literature makes our work valuable to researchers interested in entering this field.
In this endeavor, we undertake a comprehensive survey of MVC methods, viewing them through the lens of self-supervised learning. Our investigation covers commonly used datasets for images and videos, self-supervised signals, and learning paradigms. Our contributions can be summarized as follows:

\begin{itemize}
    \item We examine various data patterns with distinct properties, presenting relevant data formats and widely used image and video datasets in extensive use by researchers.

    \item We focus on multi-view clustering in self-supervised learning scenarios and review the problems and challenges due to data incompleteness over the past few years.
    
    \item We delve into self-supervised signal forms derived from representation learning methods and extend the consistency learning paradigm from self-supervised learning techniques. This chosen presentation format aids readers in comprehending the distinctions among existing methods.
    
    \item Our work offers researchers insights into prospective research directions and associated challenges.
\end{itemize}

The overall structure of the paper is as follows: 
In Section \ref{sec1}, we introduced the background and principles of self-supervised MVC and provided an overview of this paper.
Additionally, we introduce some publicly available multi-view datasets, including image and video representations, and summarize the directions and challenges derived from the incompleteness of the data in Section \ref{sec2}.
Next, we proceed to present the representation learning that can acquire self-supervised signals in Section \ref{sec3}, sub-categorized mainly by their learning methods.
In Section \ref{sec4}, we present concrete self-supervised MVC methods with generative and contrastive proxy tasks, respectively. Finally, in Section \ref{sec5} and Section \ref{sec6}, we conclude this paper and discuss challenges and future trends for self-supervised MVC. The main structure is also illustrated in Fig. \ref{fig2}.

\section{Multi-view Data and Data-related Issues}
\label{sec2}
Vision serves as a crucial channel for humans to gather external information, and different types of visual data possess distinct properties and roles, all of which are vital for achieving self-supervised MVC. In this section, as illustrated in Table \ref{table1}, we compile publicly available image and video datasets within the MVC domain, providing essential information such as their scope, format, and other relevant details. It's worth noting that a significant portion of image multi-view datasets is derived from the original single-view datasets. 
Therefore, many researchers generate various new datasets on top of these datasets on their own for further research. In addition, we also discuss that these datasets bring in new assumptions to fit more complex realities by dealing with missing and unaligned cases in simulated realities, which also creates the division of MVC from the perspective of practical problems.

\subsection{Image Dataset}
In the realm of MVC, images represent the most traditional and widely utilized form of visual data. With the advent of the big data era, there is a growing trend of employing numerous image datasets for training MVC models. In the subsequent sections, we will introduce some of the prominent image datasets individually.
\begin{table*}[]
\renewcommand\arraystretch{0.9}
\centering
\caption{Dataset summary. '-' indicates that the division of the number of categories in reality will vary with the purpose of the researcher's analysis.}
\label{table1}
\setlength{\tabcolsep}{2.1mm}{
\begin{tabular}{ccccl}
\hline
Datasets                                                                           & Type  & Size   & \# of categories & \multicolumn{1}{c}{link}                                                                                                        \\ \hline
Sence-15\cite{43}                                                                            & Image & 4485   & 15               & \href{https://figshare.com/articles/dataset/15-Scene/_Image/_Dataset/7007177/}{\begin{tabular}[c]{@{}l@{}}https://figshare.com/articles/dataset/ \\
15-Scene\_Image\_Dataset/7007177\end{tabular}  }          \\
NoisyMNIST  \cite{44}                                                                        & Image & 20k    & 10               & \href{http://yann.lecun.com/exdb/mnist/}{http://yann.lecun.com/exdb/mnist/}                                                                                         \\
Caltech101 \cite{45}                                                                         & Image & 9144   & 102              & \href{https://data.caltech.edu/records/mzrjq-6wc02}{https://data.caltech.edu/records/mzrjq-6wc02}                                                                                    \\
LandUse-21  \cite{47}                                                                         & Image & 2100   & 21               & \href{http://weegee.vision.ucmerced.edu/datasets/landuse.html}{http://weegee.vision.ucmerced.edu/datasets/landuse.html}                                                                      \\
\begin{tabular}[c]{@{}c@{}}UWA3D Multi-view\\  Activity (UWA)\end{tabular}  \cite{49}         & Image & 660    & 11               & \href{https://ieee-dataport.org/documents/uwa-3d-multiview-activity-ii-dataset}{\begin{tabular}[c]{@{}l@{}}https://ieee-dataport.org/documents/\\ uwa-3d-multiview-activity-ii-dataset\end{tabular}}             \\
\begin{tabular}[c]{@{}c@{}}Depth-included Human\\  Action dataset (DHA)\end{tabular}\cite{50} & Image & 660    & 11               & \href{https://www.researchgate.net/figure/Action-snaps-in-the-DHA-dataset_fig2_300700290}{\begin{tabular}[c]{@{}l@{}}https://www.researchgate.net/figure/\\ Action-snaps-in-the-DHA-dataset\_fig2\_300700290\end{tabular}} \\
MNIST-USPS \cite{51}                                                                         & Image & 5000   & 10               & \href{https://git-disl.github.io/GTDLBench/datasets/usps_dataset/}{https://git-disl.github.io/GTDLBench/datasets/usps\_dataset/ }                                                                   \\
BDGP  \cite{52}                                                                              & Image & 2500   & 5                & \href{https://www.fruitfly.org/}{https://www.fruitfly.org/}                                                                                                   \\
Handwritten   \cite{53}                                                                      & Image & 2000   & 10               & \href{http://yann.lecun.com/exdb/mnist/}{http://yann.lecun.com/exdb/mnist/ }                                                                                              \\
NUS-WIDE     \cite{54}                                                                       & Image & 30000  & 31               &
\href{https://lms.comp.nus.edu.sg/wp-content/uploads/2019/research/nuswide/NUS-WIDE.html}{
\begin{tabular}[c]{@{}l@{}}https://lms.comp.nus.edu.sg/wp-content/\\ uploads/2019/research/nuswide/NUS-WIDE.html\end{tabular}}   \\
Fashion     \cite{103}                                                                        & Image & 60000  & 10               & 
\href{https://github.com/zalandoresearch/fashion-mnist/blob/master/README.zh-CN.md}{
\begin{tabular}[c]{@{}l@{}}https://github.com/zalandoresearch/\\ fashion-mnist/blob/master/README.zh-CN.md\end{tabular}}         \\
SUN RGB-D    \cite{104}                                                                       & Image & 10335  & 700+             & \href{https://rgbd.cs.princeton.edu/}{https://rgbd.cs.princeton.edu/ }                                                                                                 \\
Wikipedia   \cite{105}                                                                        & Image & 2866   &     -   & \href{https://huggingface.co/datasets/wikipedia}{https://huggingface.co/datasets/wikipedia}                                                                                       \\
COIL-20    \cite{106}                                                                         & Image & 1440   & 20               & \href{https://www.cs.columbia.edu/CAVE/software/softlib/coil-20.php}{ \begin{tabular}[c]{@{}l@{}}https://www.cs.columbia.edu/CAVE/\\ software/softlib/coil-20.php\end{tabular}  }                      \\
Consumer Video(CCV)   \cite{55}                                                              & Video & 6773   & 20               & \href{https://www.ee.columbia.edu/ln/dvmm/CCV/}{https://www.ee.columbia.edu/ln/dvmm/CCV/}                                                                                        \\
YouTube Video         \cite{56}                                                              & Video & 120000 & 31               & \href{https://research.google.com/youtube8m/}{https://research.google.com/youtube8m/}                                                                                         \\ \hline
\end{tabular}
}
\end{table*}

\textbf{Sence-15} \cite{43} contains 4485 images from 15 different interior and outdoor scene categories, as well as PHOG and GIST functionality. It is widely used for various computer vision research problems, such as multi-view clustering. In terms of multi-view properties, scene 15 has a 20-dim GIST feature and a 59-dim PHOG feature that are utilized as two distinct viewpoints.

\textbf{NoisyMNIST} \cite{44} uses view 1 of the original 70k MNIST images and view 2 of white Gaussian-noise within-class images chosen at random. Since most baselines cannot handle such a large dataset in experiments, a 20k MNIST subset consisting of 10k validation images and 10k tests is usually used. 

\textbf{Caltech101} \cite{45} consists of 9,144 images distributed over 102 categories, and it has two features, i.e., the 1,984-dim HOG feature and the 512-dim GIST feature, which are extracted as two views. These images include animals, wheels, flowers, etc., and come from the same category with a very large variation of shapes. Furthermore, Caltech101-20 \cite{46} dataset, which includes 2,386 images of 20 subjects with two handcrafted features as two views, is commonly used in experiments with HOG and GIST features as two views. 

\textbf{LandUse-21} \cite{47} is a 21-class land use image dataset meant for research purposes. There are 100 images for each of the following classes, and each image measures 256x256 pixels. The images were manually extracted from large images from the USGS National Map Urban Area Imagery collection for various urban areas around the country. The pixel resolution of this public domain imagery is 1 foot. It consists of 2100 satellite images from 21 categories, with 100 images each, with PHOG and LBP features.

\textbf{UWA} (UWA3D Multi-view Activity) \cite{49} is collected by Kinect sensors with RGB and depth features. It consists of 660 action sequences, or 11 acts carried out by 12 subjects five times each. 

\textbf{DHA} (Depth-included Human Action dataset) \cite{50} consists of 660 action sequences, or 11 acts carried out by 12 subjects five times each.

\textbf{MNIST-USPS} \cite{51} is a popular handwritten digit dataset, which contains 5,000 samples with two different styles of digital images. The USPS picture is 256 dim, while the MNIST image is 784 dim. 

\textbf{BDGP} \cite{52} was used for gene expression analysis by the Berkeley Drosophila Genome Project. It is composed of 5 categories and 2500 samples, where each class has 500 samples, each of which is represented by visual and textual features. Specifically, each of these is represented by visual and textual features.

\textbf{Handwritten} \cite{53} contains five views and 2000 samples from ten numerals (i.e., 09), where the five views are obtained by Fourier coefficients, profile correlations, the Karhunen-Love coefficient, Zernike moments, and pixel average extractors.

\textbf{NUS-WIDE} \cite{54} is a real-world web image dataset collected by researchers at the National University of Singapore. In their experiments, the researchers typically use a multi-view subset containing 30,000 images and 31 classes, where each image is represented by five low-level features, i.e., color histogram, color correlation map, edge orientation histogram, wavelet texture, and chunked color moments.

\textbf{Fashion} \cite{103} has a total of 10 categories. 60,000 images and labels of 28*28 pixel points of clothes and pants are provided for training, and 10,000 images and labels of 28*28 pixel points of clothes and pants are provided for testing.

\textbf{SUN RGB-D} \cite{104} has 10,335 RGB-D images. The features are extracted from the original images using the deep neural network.

\textbf{Wikipedia} \cite{105} contains 2866 multimedia documents, which were collected from Wikipedia. Each document contains two views, i.e., the image view and the text view.

\textbf{COIL-20} \cite{106} is composed of 1,440 images of 20 objects in which the background has been discarded. Each image is represented by three kinds of features, including a 1024-dimension intensity, a 3304-dimension local binary pattern and a 6750-dimension Gabor.

\subsection{Video Datastet}
Video content, as a sequence of images presented in a temporal framework, has gained increasing prominence in comparison to static images. Platforms like YouTube and TikTok are clear examples of the growing utilization of videos by people. Videos offer a richer source of information, capturing more detailed features than static images. Learning from multi-view videos allows the acquired representations to effectively decompose functional properties, even as viewpoints and agents remain consistent. This, in turn, facilitates the subsequent learning of new tasks. Consequently, within the MVC field, researchers have shifted their focus towards using video datasets to train MVC models, aiming to attain superior results. In the following, we will introduce some of the main video datasets separately.

\textbf{CCV} (Consumer Video) \cite{55} is a video dataset with 6,773 samples belonging to 20 classes and provides hand-crafted Bag-of-Words representations of three views, such as STIP, SIFT, and MFCC.

\textbf{YouTube Video}  \cite{56} is about 120,000 films' worth of feature values, and class labels make up the dataset. Up to 13 different feature kinds, including auditory, textual, and visual features, from three high-level feature families can be used to describe each video. There are 31 class labels, 30 of which correspond to well-known video games, and the remaining nine are for different games. The data's high-quality feature representation makes it extremely useful for grouping online videos.

\subsection{Data-related Issues}
\begin{figure}[t!]
\centering
\includegraphics[width=1\linewidth]{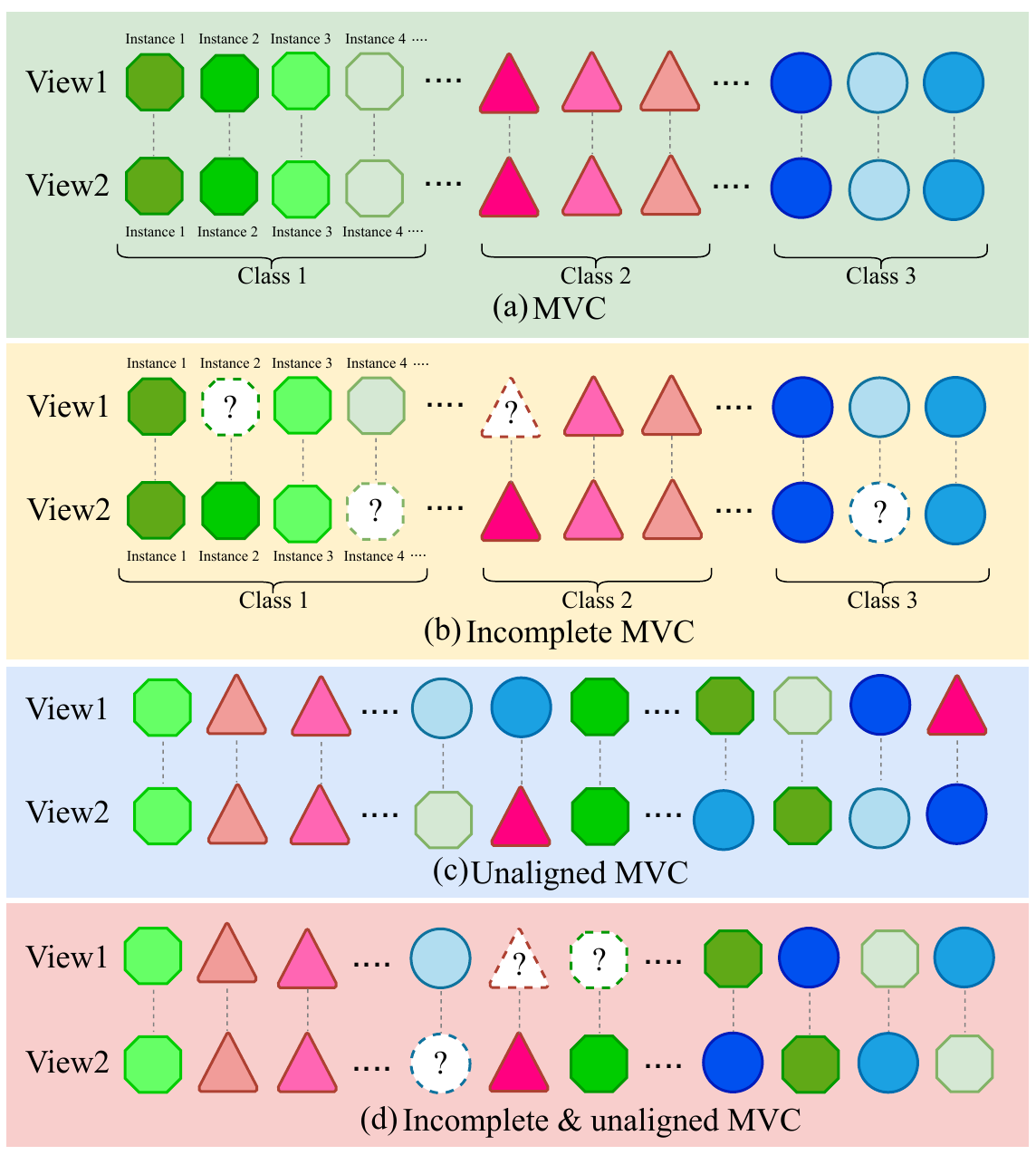}
\caption{Illustrative examples of the related data issues. Taking a bi-view data as a showcase, we use two rows of polygons to denote two views, where each column of polygons represents a pair of instances that may be incomplete or unaligned as shown in (a). Polygons with the same shape belong to one category, and the same color is a pair of aligned instances. The "?" denotes that the view sample is missing. (a) MVC: multi-view clustering based on complete and aligned multi-view data. (b) Incomplete MVC: multi-view clustering based on incomplete and aligned multi-view data. (c) Unaligned MVC: multi-view clustering based on complete and unaligned multi-view data. (d) Unaligned MVC: multi-view clustering based on incomplete and unaligned multi-view data.}
\label{fig3}
\end{figure}
Although images and videos are two different types of datasets, existing self-supervised MVC methods do not treat the two data forms any differently, as both essentially satisfy the multimodal input of the algorithm. The success of existing multi-view clustering methods \cite{1, 2, 92, 84, 81} heavily relies on the assumption of view instance completeness and consistency, referred to as complete information. However, these two assumptions would be inevitably violated in data collection and transmission, thus leading to incomplete and unaligned MVC. Researchers have therefore begun to artificially make datasets missing or unaligned to simulate more complex real-world situations \cite{5, 123, 124, 126}. This has greatly expanded the research space, and many excellent works promoting the use of self-supervised learning in MVC have emerged. Different views exhibit consistency and complementary properties of the same data, leading to extensive research on multi-view learning. As shown in Fig. \ref{fig3}, the presentation of data provides another way to divide the way we look at self-supervised MVC into the following four categories:

\textbf{MVC} aims to harness information from multiple views to enhance clustering. In the literature, existing MVC methods can be broadly categorized into two groups: traditional methods and deep methods. Traditional MVC often relies on machine learning algorithms like matrix factorization, graph learning, and kernel learning. However, these methods face challenges when applied to large-scale datasets and exhibit limited generalization capabilities. In contrast, deep MVC methods have gained popularity recently due to their exceptional representation capabilities, as acknowledged within the community \cite{14, 20, 21, 66, 75}. The majority of self-supervised MVC methods, which also fall under the deep MVC category, have emerged as the future research mainstream, largely propelled by the widespread adoption of contrastive learning in self-supervised learning.
Irrespective of the specific method employed, MVC \cite{126, 1, 2} typically necessitates access to complete and aligned multi-view data within the model to uphold consistency and accuracy. For example, Xu et al. introduced a novel framework \cite{2} for multi-level feature learning in contrastive multi-view clustering, addressing the challenge of balancing consistency and complementarity. Their Multi-VAE \cite{1}, capable of learning disentangled and interpretable visual representations while tackling large-scale data clustering tasks, has exhibited commendable performance. This is achieved by employing a generative model to regulate mutual information capacity. Consequently, generative or contrastive methods, falling under the realm of self-supervision, constitute the primary means for learning consistent information and managing complementary information in contemporary MVC research.

\textbf{Incomplete MVC} stands as a significant unsupervised method designed to cluster multi-view data that contains missing information in certain views. Initially, addressing such challenges involved simply populating the missing elements with mean average eigenvalues or other matrix representations. However, these basic padding strategies prove inadequate when the missing data rate is high. Furthermore, these methods struggle to fully capture the latent information within the missing data, as the missing elements are not adequately recovered.
In recent times, as self-supervised learning has evolved, generative models such as Generative Adversarial Network (GAN) \cite{101} and Variational Autoencoder (VAE) \cite{114} have demonstrated superior performance, particularly in scenarios where significant data is available for training. For instance, Wang et al. introduced a deep Incomplete Multi-View Clustering (IMC) method \cite{107} based on GAN. Lin et al. proposed a novel objective that unifies representation learning and data recovery within a cohesive framework, viewing it from the perspective of information theory. Notably, they integrated generation and contrastive learning into a unified and consistent learning framework \cite{5}.
DIMVC \cite{4}, currently recognized as the State-of-the-Art (SOTA) method, converts complementary information into a supervised signal with high confidence. Its primary objective is to establish multi-view clustering consistency for both complete and incomplete data.

\textbf{Unaligned MVC} represents a relatively new and less-explored direction within the field, with limited existing research and considerable room for development. In recent years, substantial efforts have been directed towards addressing incomplete MVC, primarily through the imputation of missing samples using various data recovery methods \cite{5, 7, 161}. In contrast to incomplete MVC, unaligned MVC is a relatively uncharted territory, emerging only recently \cite{123, 124}.
A plausible method to tackling unaligned MVC begins by realigning the data using the Hungarian algorithm \cite{122}, followed by performing MVC based on the realigned data. However, this method is unsuitable for large datasets due to the non-differentiable nature of the Hungarian algorithm. To address this limitation, PVC \cite{123} proposes the utilization of a differentiable surrogate for the non-differentiable Hungarian algorithm. This allows it to be recast as a pluggable module, and subsequently, it constructs the distance matrix to supervise alignment correspondence in the latent space.
Nevertheless, both the original Hungarian algorithm and PVC focus on achieving instance-level alignment, which may be insufficient for MVC. The core of clustering and classification lies in establishing a one-to-many mapping, thus rendering category-level alignment more advantageous. Subsequent work by MvCLN \cite{124} reframes the view alignment problem as an identification task. It introduces a novel noise-robust contrastive loss designed to mitigate or even eliminate the impact of noisy labels that may arise during pair construction.

\textbf{Incomplete \& unaligned MVC} represents a relatively unexplored problem, emerging only recently \cite{123}. This direction aligns more closely with real-world challenges. To date, the sole work addressing this issue is SURE \cite{126}, which strives to learn categorical similarities and establish correspondences across views. SURE achieves this by leveraging a novel noise-robust contrastive learning paradigm initially proposed by Yang et al.
\section{Representation Learning Method}\label{sec3}
Representation learning plays a pivotal role in self-supervised MVC and the process of generating and utilizing self-supervised signals. In this section, we provide an in-depth discussion of the various forms in which self-supervised signals manifest and categorize them into five distinct categories. In practice, the concept of self-supervised signals is exceptionally versatile, encompassing feature representations, mathematical functions, pseudo-labels, or even algorithmic designs. This versatility provides self-supervised MVC with a vast research space and practical significance.
Researchers inject prior information into representation learning algorithms or devise proxy tasks to guide the network's optimization, thereby generating self-supervised signals. Appropriately designed self-supervised signals have the capacity to harness the complementary and consensus information present across multiple views, facilitating the clustering of objects into distinct partitions \cite{8}. The consensus principle strives to maximize consistency among different views, while the complementarity principle recognizes that each view contains unique information not found in others. Consequently, gaining a deeper understanding of self-supervised signaling can offer valuable insights into representation learning and drive progress in future research endeavors.
\begin{figure}[h]
\centering
\includegraphics[width=1\linewidth]{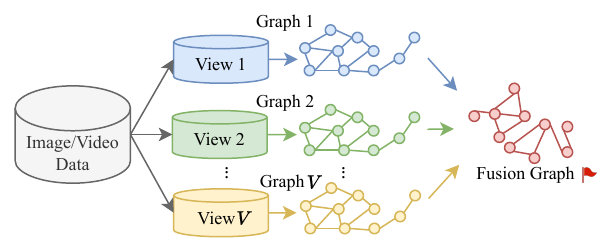}
\caption{General procedure of graph-based multi-view clustering representation learning. $V$ is the number of views.}
\label{fig4}
\end{figure}
\subsection {Graph-based Representation}
Graphs are widely used to represent the relationships between different samples, where nodes represent data samples and edges represent the relationships between samples. This graphical method offers a more comprehensive means of encapsulating the richness of multi-view data. To unearth the underlying clustering structure more effectively, it's crucial to efficiently integrate these graphs in a mutually reinforcing manner.
In essence, self-supervised signals based on graph representations are derived through the fusion of multi-view data, and the generalized process aligns with the illustration in Fig. \ref{fig4}. Specifically, we can formulate the generic problem of multi-view graph clustering as follows:
\begin{equation}
\label{eq1}
\begin{aligned}
\min _{\mathbf{A}^{(v)}, \mathbf{S}} \sum_{v=1}^{V} \sum_{i \neq j}^{N} \operatorname{Dis}\left(\mathbf{x}_{i}^{(v)}, \mathbf{x}_{j}^{(v)}\right) \mathbf{A}^{(v)}+\lambda \Psi\left(\mathbf{S}, \mathbf{A}^{(v)}\right) \\
\text { s.t. } \quad \mathbf{A}^{(v)} \geqslant 0, \mathbf{a}_{i}^{(v)} \mathbf{1}=1, \mathbf{S} \geqslant 0, \mathbf{s}_{i} \mathbf{1}=1
\end{aligned}
\end{equation}
where $\mathbf{x}_{i}^{(v)}$ represents the $i$-th feature vector for the $v$-th view, ${A}^{(v)}$ is the affinity map for a particular view, and $\mathbf{S}$ is the consistent similarity matrix of multiple affinity graphs after some fusion. $Dis(.)$ denotes some similarity or distance metric (e.g., Euclidean distance).
$\Psi(.)$ is certain fusion function to combine multiple $\mathbf{A}^{(v)}$ to obtain the final consensus cluster assignment $\mathbf{S}$.

The graph convolutional network (GCN)-based model \cite{78, 79, 80, 81, 82} employs a deep embedding method for MVC and incorporates graph structure information into its model. This allows for the utilization of both the information from the graph structure and the node features.
Furthermore, self-supervised GCN constructs a new view descriptor tailored for graph-structured data. Simultaneously, the generated self-supervised signals guide the learning process for latent representations and coefficient matrices. These learned representations and matrices are subsequently used to perform node clustering.
Among the early works in this domain, O2MA \cite{78} introduced graph autoencoders for learning node embeddings based on a single information graph. It employs the captured view consistency in low-dimensional feature representations as a fine-tuning stage for self-supervised signals.
Subsequently, in the following years, numerous works \cite{79, 80, 81, 82} on self-supervised MVC based on graph representations emerged:
1) Cheng et al. developed a model that employs two-pathway encoders to map graph embedding features and learn view-consistency information. This method explores graph embeddings and consistent embeddings of high-dimensional samples \cite{79}.
2) MDGRL \cite{80} is built upon the concept of the graph autoencoder for local feature learning and is a valid variant of the variational graph autoencoder for global deep graph representation learning.
3) Cai et al. combined global and partial GCN autoencoders to create a self-training clustering module with adaptively weighted fusion. They use this module to simultaneously mine the global and unique structures from various viewpoints \cite{81}.
4) Xia et al. imposed the diagonal constraint on the consensus representation that generated by multiple GCN autoencoders with the self-supervised clustering scheme better clustering capability \cite{82}. They utilize the clustering labels produced to supervise the self-expressive coefficient matrix $\mathbf{C
}$, specifically,
\begin{equation}
\begin{aligned}
\min _{\mathbf{C}} \sum_{i, j=0}^{n}\left|\mathbf{c}_{i j}\right| \frac{\left\|\widehat{\mathbf{l}_{i}}-\widehat{\mathbf{l}_{j}}\right\|_{2}^{2}}{2},
\end{aligned}
\end{equation}
where $\widehat{\mathbf{l}}_{i}, \widehat{\mathbf{l}}_{j} \in \widehat{\mathbf{L}}$ represent the label vector corresponding to the $i$-th and $j$-th nodes, respectively.
5) In the latest work, DMVCJ \cite{24} utilizes the latent graphs to promote the performance of deep-embedded MVC models from two aspects: the global weights act as self-supervised signals and also mitigate the noise problem.

As contrastive learning continues to evolve, several methods \cite{18, 19, 24, 67, 68} have emerged as self-supervised methods for learning node and graph-level representations through multi-view contrastive graph clustering.
Hassani et al. introduced a self-supervised method for learning node and graph-level representations by contrasting structural views of graphs \cite{19}. However, they overlook the fact that the original graph data may contain noise or be incomplete, rendering their method less directly applicable.
To address this challenge, two subsequent works have been proposed. First, Pan et al. employ graph filtering techniques to filter out undesirable high-frequency noise while preserving the essential geometric features of the graph. This results in a smoother representation of nodes, which is then used to learn a consensus graph regularized by a graph contrastive loss \cite{68}.
Second, SGCMC \cite{18} utilizes clustering labels to guide the learning of latent representations and coefficient matrices. These learned representations and matrices are subsequently employed for node clustering. SGCMC constructs a new view descriptor for graph-structured data by mapping the original node contents to a complex space using the Euler transform. This method not only suppresses outliers but also unveils the nonlinear patterns within the embedded data.
In addition to the challenge of sample noise, the problem of sample missing is more practical. ACTIVE \cite{67} is designed with both intra-view graph contrastive learning and cross-view graph consistency learning to maximize the mutual information across different views within a cluster.
Graphs being a discrete data structure often exhibit tight correlations in common graph learning tasks. Consequently, the design of graph contrastive learning algorithms tailored to these properties and how contrastive learning can effectively enhance graph representation and node representation continue to be active areas of exploration.

\subsection {Subspace-based Representation}
\begin{figure}[h]
\centering
\includegraphics[width=1\linewidth]{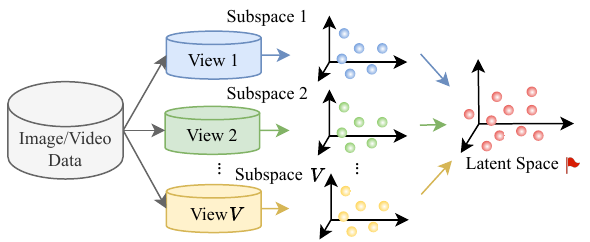}
\caption{General procedure of subspace-based multi-view clustering representation learning.}
\label{fig5}
\end{figure}
Multi-view subspace clustering methods usually either learn a shared and unified subspace representation from multiple view-specific subspaces or discover a latent space for high-dimensional multi-view data to reduce the dimensionality, based on which latter subspace learning is conducted. To illustrate, the general process of the multi-view subspace representation learning method is shown in Fig. \ref{fig5}, where the subspace representation itself is usually used as a self-supervised signal. In addition, the objective formula for multi-view subspace clustering can be rewritten
\begin{equation}
\begin{aligned}
\min _{\mathbf{z}^{(v)}} \sum_{v=1}^{m}\left\|\mathbf{x}^{(v)}-\mathbf{x}^{(v)} \mathbf{z}^{(v)}\right\|_{F}^{2}+\lambda \Omega\left(\mathbf{z}^{(v)}\right)+\\
\gamma \Psi\left(\mathbf{z}, \mathbf{z}^{(v)}\right) \\
\text { s.t. } \quad \mathbf{z}^{(v)} \geqslant 0, \mathbf{z}^{(v)} \mathbf{1}=1
\label{con:inventoryflow}
\end{aligned}
\end{equation}
where $\mathbf{x}^{(v)}$ denotes the data of the $v$-th view, ${\mathbf{z}}^{(v)}$ denotes the view-specific subspace representation, and $\mathbf{z}$ is the learned common representation. $ \Omega(.)$ stands for the certain regularization term about ${\mathbf{z}}^{(v)}$. To discover the latent space, there is another objective function which can be formulated as
\begin{equation}
\begin{aligned}
\min _{\mathbf{z}} \sum_{v=1}^{m} \mathcal{F}\left(\mathbf{x}^{(v)}, \mathbf{H}\right)+\lambda\|\mathbf{H}-\mathbf{H} \mathbf{z}\|_{F}^{2}, \\
\text { s.t. } \quad \mathbf{z} \geqslant 0, \mathbf{z} \mathbf{1}=1,
\end{aligned}
\end{equation}
where $\mathbf{H}$ denotes the latent space learned from multiple views. Then subspace learning is performed on the basis of consensus $\mathbf{H}$.

In recent times, there has been a growing emergence of deep learning-based methods for multi-view subspace clustering. These methods aim to enhance the learning of representations for each view and uncover common latent subspaces \cite{14, 15, 21, 26, 75, 94}.
Drawing inspiration from \cite{74}, Abavisani et al. introduced a self-expressive layer designed to enforce the self-expressiveness property. This addition contributed to advancements in subspace reconstruction \cite{73}.
Following this, Cui et al. proposed SG-DMSC \cite{75}, which introduced a novel loss term called spectral supervision. This addition simplifies the consensus clustering process, resulting in improved clustering performance. However, prior work often treated spectral clustering and affinity learning as separate entities.
Sun et al. introduced S2DMVSC \cite{14}, a framework that seamlessly integrates spectral clustering and affinity learning within a deep learning context. It leverages clustering results to guide latent representation learning for each view and common latent subspace learning across multiple views.
DASIMSC \cite{15} proposed a dual-aligned, self-supervised, incomplete multi-view subspace clustering network. This method maintains consistency in semantics between the inherent local structure within a view and the incomplete view.
In addition, there is work focused on subspace and contrast learning. The SCMC method \cite{26} utilizes view-specific autoencoders to map raw multi-view data into compact features, capturing their nonlinear structure. Subsequently, subspace learning is employed to unify multi-view data into a shared semantic space.
As research in this field deepens, the challenge of information bottleneck has become increasingly prominent. Addressing this issue is crucial for performance enhancement. SIB-MSC \cite{21}, as the pioneering work exploring information bottlenecks in multi-view subspace clustering, learns minimal sufficient latent representations for each view guided by self-supervised information bottleneck principles. This method helps uncover common information shared across different views.

Recently, unified representations of latent space were fed into an off-the-shelf deep clustering model in order to produce the clustering results. Due to the powerful representation capability of deep learning, multi-view subspace clustering has achieved good performance on deep multi-view subspace clustering networks, self-supervised multi-view deep subspace clustering networks, generalized latent multi-view subspace clustering, and other methods.
\begin{figure}[h]
\centering
\includegraphics[width=1\linewidth]{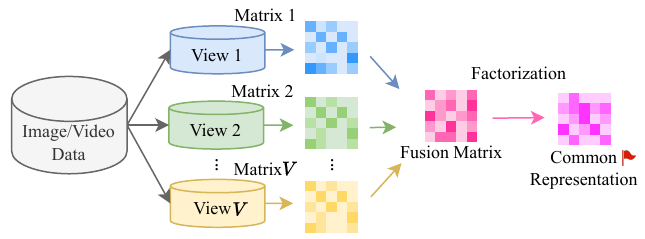}
\caption{General procedure of matrix factorization -based multi-view clustering representation learning.}
\label{fig6}
\end{figure}
\subsection {Matrix-based Representation}
A fundamental assumption in multi-view data analysis is the existence of a consistent label distribution across different views, often referred to as multi-view semantic consistency. When employing matrix factorization for multi-view clustering, the primary goal is to decompose the multi-view data into consensus representations that are shared by all views. This decomposition process aims to generate self-supervised signals capable of guiding and training subsequent learning stages.
The hierarchical decomposition of the dataset $X$ by the depth model can be expressed as
\begin{equation}
\begin{aligned}
X \approx F_{1} F_{2} \cdots F_{m} R_{m}
\end{aligned}
\end{equation}
where $F_1,F_2,...,F_m$ denotes a series of mapping matrices, $m$ denotes the total number of layers and $R_m$ denotes the final common latent representation. A generic representation learning process for matrix-based decomposition is shown in Fig. \ref{fig6}.

Many studies \cite{65, 66, 76, 85} focusing on multi-view matrix decomposition have yielded impressive results.
Furthermore, within the framework of multi-view matrix decomposition, we systematically break down less important factors layer by layer, ultimately generating an effective consensus representation in the final layer of the MVC process. This method has significantly enhanced clustering performance.
Zhao et al. \cite{65} introduced a deep matrix decomposition framework tailored for MVC. They employed semi-negative matrix decomposition to hierarchically learn common feature representations with greater consistency across views. This method ensures that the consensus representation retains most of the shared structural information across multiple graphs. Notably, this was the pioneering attempt to apply semi-negative matrix decomposition to self-supervised MVC.
For the sake of simplicity, the SMDMF method \cite{66} automatically assigns weights to each view without the introduction of additional parameters. This method is capable of autonomously assigning suitable weights to each view for information fusion, enabling the creation of a shared matrix representation without the need for additional hyperparameters. 
Recently, Wei et al. introduced DMClusts \cite{76}, a method designed to identify multiple clusters within multi-view data. DMClusts employs a progressive method by decomposing the multi-view data matrix into a layer-by-layer representation subspace, generating a cluster at each layer. This innovative method leverages deep matrix decomposition in conjunction with deep learning techniques and introduces a novel metric, balanced diversity, to uncover multiple distinct and high-quality clusters.
In summary, while matrix decomposition is a well-established machine learning technique, its integration with deep learning has significantly enhanced algorithm efficiency. Nevertheless, it's worth noting that due to the complexity of algorithm design and the computational resources required, its application in self-supervised MVC is somewhat limited compared to other representation learning methods.
\begin{figure}[h]
\centering
\includegraphics[width=1\linewidth]{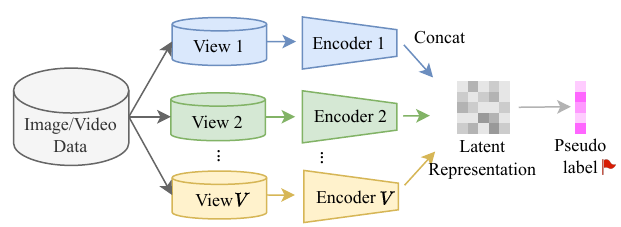}
\caption{General procedure of pseudo label-based multi-view clustering representation learning.}
\label{fig7}
\end{figure}
\subsection {Pseudo label-based Representation}
Researchers have made a significant discovery indicating that learning a unified latent representation of multi-view data through pseudolabeling can greatly enhance clustering performance \cite{88, 89, 90}. The use of pseudo-label constraints helps retain the view-specific characteristics of multi-view data while learning a unified latent representation. This method simultaneously preserves internal view similarities and inter-view relationships. As illustrated in Fig. \ref{fig7}, the typical process involves extracting latent representations using view-specific encoders and subsequently combining these latent representations from all views to generate pseudo-labels through the K-means algorithm.

Generating accurate pseudo labels is of paramount importance, given their inherent high confidence level. SG-DMSC \cite{75}, for instance, utilized spectral clustering to generate pseudo labels as a self-guided multi-view encoder fusion layer, thereby exploring features conducive to clustering and obtaining improved latent representations. They also developed a theoretically derived alternative iterative optimization algorithm to rationalize the pseudo labels effectively.
To harness the learned pseudo-labels to their fullest potential, Kheirandishfard et al. \cite{74} adopted a self-supervised strategy to construct the objective function. Furthermore, methods that incorporate pseudolabeling as a self-supervised signal often combine it with other representation learning methods. For example, L-MSC \cite{97} and PLCMF \cite{85} integrate matrix decomposition to enhance the consistency between the affinity matrix and the learning assignment matrix. These methods are designed with self-iterating modules that impose pseudolabeling constraints on top of them.
However, some of the aforementioned work overlooks the issue of highly ambiguous clustering structures. SDMVC \cite{20}, on the other hand, exploits complementary information to construct global features, which leads to more accurate pseudo labels. These labels are then employed to learn more discriminative features and achieve consistent predictions across multiple views.
Overall, multi-view clustering methods guided by pseudolabels have demonstrated excellent performance \cite{74, 75, 85, 93, 94}. This method addresses the scalability challenges posed by existing methods when dealing with large-scale datasets and ensures a more integrated and correlated learning process across the two stages.
\begin{figure}[h]
\centering
\includegraphics[width=1\linewidth]{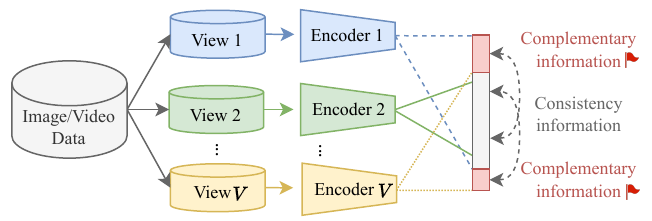}
\caption{General procedure of Complementary information-based multi-view clustering representation learning.}
\label{fig8}
\end{figure}
\subsection {Complementary information -based Representation Representation}
As illustrated in Fig. \ref{fig8}, different views often contain complementary information. To describe data objects more comprehensively and accurately, it becomes necessary to leverage this complementary information for enhanced internal clustering across multiple views, thereby providing deeper supervisory signals \cite{4}.
Complementary information present in multiple views can be harnessed to improve the performance of multi-view clustering. By combining information from multiple views, a more comprehensive representation of the target object can be obtained. In essence, multi-view clustering integrates the acquired representation features through the combination of view-specific depth encoders and graph embedding strategies within a unified framework. This method captures both high-level features and local structures from each view, effectively amalgamating the strengths of each view while mitigating their respective shortcomings.

SG-DMSC \cite{75} introduces a view-fusion layer to exploit complementarity across multiple views, but its integration is relatively straightforward and lacks depth in information comprehensiveness and structural representation.
CDIMC-net \cite{6}, on the other hand, incorporates high-level features and local structures from each view by combining view-specific depth encoders and graph embedding strategies within its framework, resulting in more effective fusion representation.
Furthermore, the concept of complementarity has been extended to address data deficiencies in methods like GP-MVC \cite{95} and DIMVC \cite{4}. GP-MVC \cite{95} implements a weighted adaptive fusion method to leverage complementary information among different views, with the learned common representation aiding in data imputation.
These methods either mine complementarity by fusing multiple similarity matrices or employ fusion layers. While fusion layers can be effective, some views may have a negative impact on the fusion process due to their inherently low quality or inaccurate estimations \cite{6,95}.
DIMVC \cite{4}, however, takes a different method by implementing a high-dimensional mapping that linearly transforms linearly separable clustering information into complementary information. This information is used as high-confidence supervised data to ensure consistent clustering assignments across all views, even for incomplete data. By concatenating embedded features from all views to create global features, DIMVC overcomes the negative impact of unclear clustering structures in certain views.
In summary, existing multi-view clustering methods typically explore complementarity between multi-view data through a fusion process \cite{4,6,75,95,96}.

\section{Self-supervised Learning Method}\label{sec4}
Generation and contrastive methods are two of the most crucial techniques in self-supervised learning extensively applied in the field of multi-view clustering, where they have made significant advancements. Generative methods aim to grasp the underlying data distribution and employ generative models to represent the data. Contrastive methods, on the other hand, directly optimize an objective function that involves pairwise similarities to minimize the average similarity within clusters and maximize the average similarity between clusters. More specifically, self-supervised MVC leverages the input data itself as supervision, effectively extracting transformation and relational information from the data across various perspectives. In this section, we categorize self-supervised MVC models into four groups: generative, contrastive, generative-contrastive, and other, with specific subcategories within each.

\subsection{Generative Methods}
In the context of multi-view clustering, the self-supervised generative method involves utilizing the inherent data structure across multiple views to learn representations that improve clustering performance without relying on external labels. This method often involves three common types of generation models in MVC:
1) AE(Autoencoder) is used to directly synthesize decoded data. 2) GAN employs a competitive process to generate missing data while obtaining backpropagation signals to refine the generation process. 3) VAE is used to learn interpretable representations in self-supervised MVC. 
These generation processes are typically embedded within the data reconstruction process of multi-view data, making autoencoders an essential component. In the following subsections, I will describe multi-view learning works for clustering tasks categorized by specific data modality combinations in the form of "AE+X," where X represents the generation method.

\subsubsection{Autoencoder}
\begin{figure}[h]
\centering
\includegraphics[width=1\linewidth]{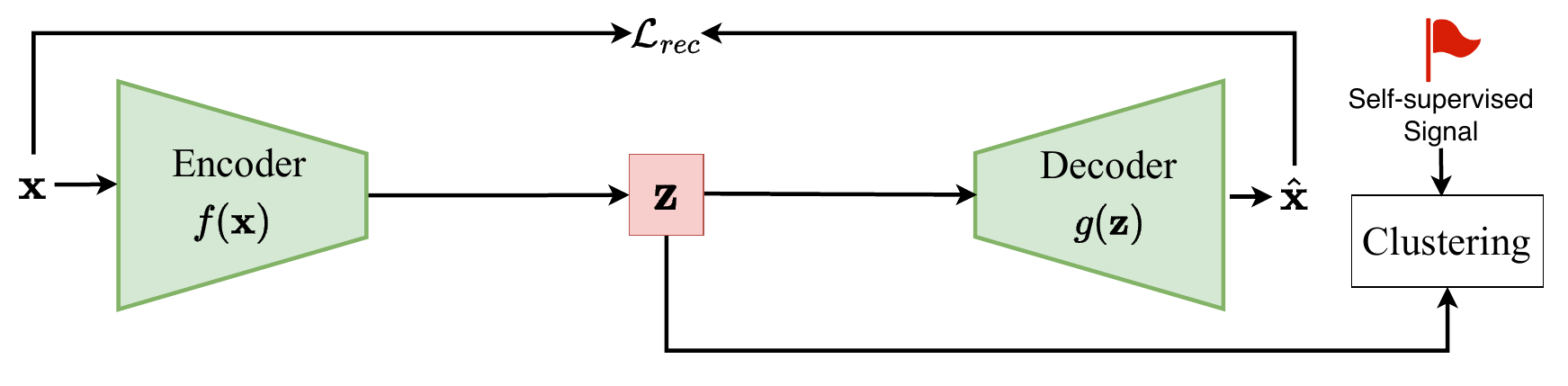}
\caption{Self-supervised multi-view clustering based on autoencoders. $f$ and $g$ denote the encoder and decoder respectively, $\mathbf{x}$ is the data of a particular view and $\hat{\mathbf{x}}$ is the decoded representation. $\hat{\mathbf{x}}$ and $\mathbf{x}$ have to be as similar as possible, so the loss function $\mathcal{L}_{r e c}=\left\|\mathbf{x}-\hat{\mathbf{x}}\right\|_{2}^{2}$ is as small as possible. Finally, clustering is accomplished based on the latent representation $\mathbf{z}$ with self-supervised signals.}
\label{fig9}
\end{figure}

An autoencoder consists of two main components: an encoder and a decoder \cite{129}, as shown in Fig. \ref{fig9}. The Encoder network captures the most salient features from the high-dimensional data and the decoder network aims to recover the data from the encoded features. So almost all MVC methods are extensions of the autoencoder. Specifically, for a sample $\mathbf{x}$, the activity value of the intermediate hidden layer of the autoencoder is the encoding of $\mathbf{x}$, mathematically:
\begin{equation}
\mathbf{z}=f\left(\mathbf{w}^{(1)} \mathbf{x}+\mathbf{b}^{(1)}\right).
\end{equation}
The output of the autoencoder is the reconstructed data:
\begin{equation}
\hat{\mathbf{x}}=g\left(\mathbf{w}^{(2)} \mathbf{z}+\mathbf{b}^{(2)}\right),
\end{equation}
where $\mathbf{w}^{(1)}$ and $\mathbf{b}^{(1)}$ are the parameters of the encoder $f$, and $\mathbf{w}^{(2)}$ and $\mathbf{b}^{(2)}$ are the parameters of the decoder $g$, and these parameters are obtained by gradient descent training. $\widehat{\mathbf{x}}$ is the result of $\mathbf{x}$ being fed to the autoencoder for reconstruction.

The autoencoder not only projects multi-view data into a latent space but also serves as a generator to produce missing views and more. AE was initially introduced in \cite{129} for pre-training artificial neural networks. Subsequently, several AE-based methods have been gradually introduced for self-supervised MVC to enhance the reconstruction of inputs from corrupted data \cite{5,70,73,74,92,123,128,132,133}. Building on these developments, Abavisani et al. introduced AE into deep multi-view subspace clustering methods for the first time in \cite{73}. Moreover, in addition to employing individual autoencoders in the aforementioned method, Xu et al. implemented a collaborative training scheme involving multiple AE networks to effectively extract complementary and consistent information from each view. This method enriched the latent representations of each view with comprehensive information \cite{70}. Furthermore, compared to the method proposed in \cite{70}, Yang et al. expanded on this method by incorporating heterogeneous graph learning, which fused the latent representations from different views using adaptive weights \cite{132}.

Cui et al. utilized spectral clustering to generate pseudo-labels, serving as self-guidance for the multi-view encoder fusion layer \cite{92}. However, this method overlooks the diversity in representations generated by the autoencoder (AE) for each view. To address this concern, Zhu et al. introduced a multi-view deep subspace clustering network (MvDSCN) in their work \cite{133}. This method learns multi-view self-representations in an end-to-end manner by integrating convolutional AEs and self-representations. MvDSCN comprises two sub-networks: the Diversity Network (Dnet) and the Universal Network (Unet). The latent space is constructed using a deep convolutional AE, utilizing fully connected layers to develop a self-representation matrix within the latent space. Specifically, the deep convolutional AE takes hand-crafted features or raw data as input and learns for each view. Moreover, Dnet learns a view-specific self-representation matrix, while Unet learns a shared self-representation matrix for all views. Alignment between each view's self-representation matrix and the common self-representation matrix is achieved through universal regularization. The AE reconstruction loss formula for MvDSCN is expressed as follows:
\begin{equation}
\min \left\{\begin{array}{c}
\left\|\mathbf{x}^{(1)}-\hat{\mathbf{x}}^{(1)}\right\|^{2}+\left\|\mathbf{x}^{(2)}-\hat{\mathbf{x}}^{(2)}\right\|^{2}+\\
...\left\|\mathbf{x}^{(V)}-\hat{\mathbf{x}}^{(V)}\right\|^{2}
\end{array}\right\},
\end{equation}
where $\mathbf{x}^{1}, \ldots, \mathbf{x}^{2}, \ldots, \mathbf{x}^{(v)}$ represent the input of multiple views, $V$ denote the number of views. 

Recently, AE have gained significant attention and found applications in addressing unaligned MVC challenges. Huang et al. introduced PVC \cite{123}, which focuses on learning specific latent spaces for each view (denoted as the $v$-th view) using autoencoders to minimize reconstruction errors. This method notably enhances clustering performance, especially when dealing with partially aligned data. However, these methods do not address solutions for handling missing data.
In contrast, Lin et al. proposed COMPLETER \cite{5}, which integrates representation learning and data recovery within a unified framework from an information-theoretic perspective. The COMPLETER includes two training modules: view-specific autoencoders and cross-view prediction networks. For each view, the COMPLETER utilizes an autoencoder to extract the latent representation $\mathbf{z}^{(v)}$ by minimizing the reconstruction loss $\mathcal{L}_{rec}$. It is worth noting that the AE structure is instrumental in preventing trivial solutions.
In summary, autoencoders offer several advantages, including the ability to learn shared low-dimensional representations from multiple views, extract latent data features, and generate new views to enhance data richness and completeness. However, they are susceptible to overfitting, and compared to other generative methods, they may face challenges in capturing complex relationships between different views, which can lead to suboptimal accuracy in the generated representations.

\subsubsection{Autoencoder + GAN}
\begin{figure}[h]
\centering
\includegraphics[width=1\linewidth]{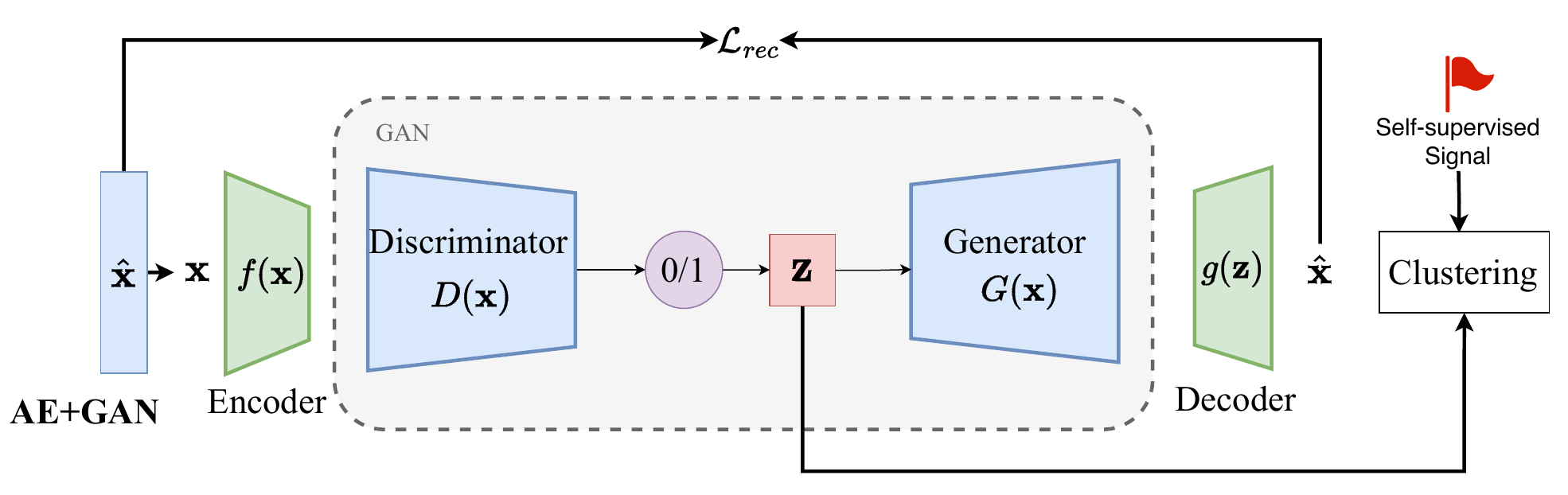}
\caption{GAN-based self-supervised multi-view clustering. $D(\mathbf{x})$ denotes the discriminator and $G(\mathbf{z})$ denotes the generator. $\mathbf{x}$ is the view-specific data, $\hat{\mathbf{x}}$ is the decoded representation, and $\mathbf{x}$ and $\hat{\mathbf{x}}$ must be as similar as possible. Finally, clustering is completed based on the latent representation $\mathbf{z}$ with self-supervised signals.}
\label{fig10}
\end{figure}

GAN is a fundamental model. Its core idea lies in the Generator, which learns the characteristics of data distribution from random noise, while the Discriminator distinguishes real data from generated data, as illustrated in Fig. \ref{fig10}. The objective of generating high-quality data is achieved through this adversarial process. Similar to the foundational concept of autoencoders, GAN, as a more innovative extension, holds a crucial role in the field of data generation. The classic GAN loss is represented by Formula \ref{18}.

Recent research has demonstrated the widespread use of GAN and their substantial impact on improving clustering performance in self-supervised MVC \cite{4,99,101,108,109,110}. In the context of MVC, two common types of GAN utilization have emerged:
1) Employing generative adversarial networks to generate missing data, as evidenced in \cite{107,95,127,100}.
2) Harnessing GANs to capture data distribution and unlock latent spaces through adversarial multi-view clustering networks, as outlined in \cite{28,128}. 
The integration of GAN into the autoencoding process not only maximizes the utility of new features from partial data but also enhances the model's robustness in scenarios with high data missing rates.

Currently, GANs have made substantial advancements in data generation, and their application in partial MVC has also been extensively investigated \cite{110,135}. Wang et al. introduced a novel and consistent Generative Adversarial Network for partial multi-view clustering in their work \cite{100}. This method is designed to learn a shared low-dimensional representation and employs a combination of GAN and deep embedding clustering layers to capture the common embedding structure within partial multi-view data. It serves a dual purpose: generating missing view data and enhancing the capture of common structures for clustering.
In contrast to other methods, this method utilizes a publicly shared representation encoded by each view to generate missing data for the corresponding view through an adversarial network. It employs both an encoder and a clustering network in the process. This method is intuitive and meaningful, as encoding the public representation and generating missing data within the model mutually benefit each other. This consistent GAN method not only captures a more robust clustering structure but also infers missing views effectively. Furthermore, the model fully exploits the complementary information present in multi-view data, distinguishing it from some GAN models that employ random noise for data generation.

Building upon the support of the aforementioned research, an increasing number of MVC studies have integrated GAN for multi-view data generation. For instance, AIMC \cite{107} combines monadic reconstruction and GAN techniques to make inferences about missing data. Additionally, Wang et al. introduced GP-MVC \cite{127}, which utilizes a GAN model to learn local view representations, capture shared clustering structures, and address missing data issues.
In the GP-MVC, a multi-view encoder network is employed to encode latent shared representations among multiple views. View-specific generative adversarial networks are designed to predict missing view data conditions based on latent representations from other views. Adversarial training is utilized to explore consistent information across all views. In this setup, the generator of GP-MVC aims to complete the missing data, while the discriminator's role is to differentiate between false and true data for each view.

In the domain of self-supervised MVC, beyond the application of GANs for generating missing data, there has been substantial research into using GAN to capture underlying data distributions and features. Li et al. introduced the DAMC network \cite{28}, which employs GAN as a regularizer to guide the encoder's training. The encoder captures the data distribution for each individual view and subsequently reveals the common latent space. However, DAMC does not address the challenge of preserving low-dimensional information embeddings in multi-view networks.
In response to this limitation, Sun et al. proposed a novel GAN framework for multi-view network embedding, named MEGAN \cite{128}. This method aims to retain information from each individual network view while considering the consistency and complementarity between different views. MEGAN leverages GANs for multi-view network embedding to tackle the key challenge of modeling not only the connections between different views but also the intricate associations between these views. It introduces a new GAN framework for learning low-dimensional embeddings that typically exhibit nonlinearity while preserving information in a given multi-view network. This method yields significant performance enhancements in the realm of self-supervised MVC. The generator models multi-view connectivity to produce synthetic samples capable of deceiving the discriminator, which, in turn, distinguishes true pairs of nodes from counterfeit pairs.

Through adversarial competition between the generator and the discriminator, GAN can acquire latent distribution features of data and generate missing data from random noise. While GAN excels at generating authentic samples, its training process is intricate, demanding a delicate balance between the performance of the generator and the discriminator. Furthermore, when compared to alternative generative methods, GAN places a strong emphasis on data generation and offers distinct advantages in data learning for self-supervised MVC.

\subsubsection{Autoencoder + VAE}
\begin{figure}[h]
\centering
\includegraphics[width=1\linewidth]{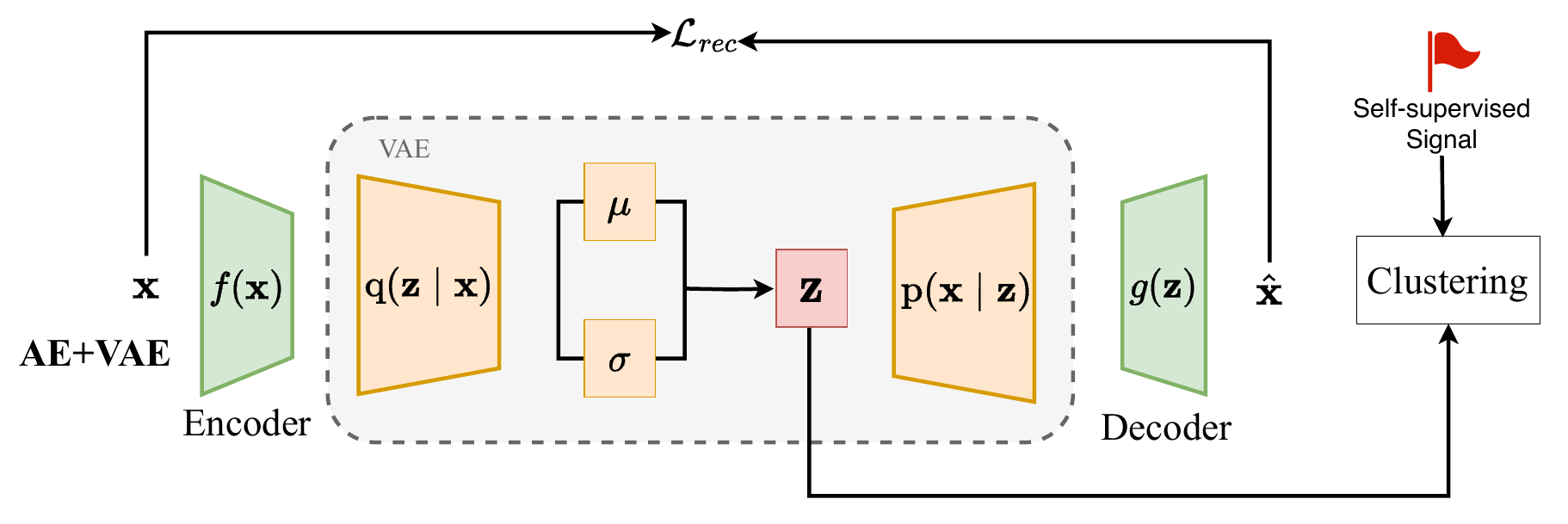}
\caption{VAE-based self-supervised multi-view clustering. The input sample $\mathbf{x}$ is obtained by the encoder $\mathrm{q(\mathbf{z} \mid \mathbf{x})}$ to obtain a vector of mean and standard deviation, and then the hidden vector $z$ is obtained by sampling, followed by the decoder $\mathrm{p(\mathbf{x} \mid \mathbf{z})}$ to obtain the output $\mathbf{x}$ and $\hat{\mathbf{x}}$ must be as similar as possible. Finally, clustering is accomplished based on the hidden variable $\mathbf{z}$ with self-supervised signals.}
\label{fig11}
\end{figure}

The fusion of variational inference and autoencoder techniques gave rise to the Variational Autoencoder (VAE) \cite{114}. In contrast to GAN, VAE entails concurrent learning of both the generator and the inference network, known as the Encoder. Its primary objective is to map data into a latent space and subsequently generate data samples from this latent space, as depicted in Fig \ref{fig11}. The objective function for VAE can be expressed as follows:
\begin{equation}
\begin{aligned}
L\left(\theta, \phi ; \mathbf{x}_{i}\right)=-D_{\mathrm{KL}}\left(q_{\phi}\left(\mathbf{z} \mid \mathbf{x}_{i}\right) \| p_{\theta}(\mathbf{z})\right)+ \\
E_{q_{\phi}\left(\mathbf{z} \mid \mathbf{x}_{i}\right)}\left[\operatorname{log} p_{\theta}\left(\mathbf{x}_{i} \mid \mathbf{z}\right)\right],
\end{aligned}
\end{equation}

$\mathbf{x}_{i}$ denotes the $i$-th sample of the input dataset, $\mathbf{z}$ denotes the latent variable, $p_{\theta}(\mathbf{z})$ denotes the prior distribution, and $q_{\phi}(\mathbf{z} \mid \mathbf{x}_{i})$ denotes the encoder generated the posterior distribution of the latent variable $\mathbf{z}$, and $p_{\theta}(\mathbf{x}_{i} \mid \mathbf{z})$ denotes the generative model.

Given the outstanding advantages of Variational Autoencoders (VAE) in feature learning and data generation, there has been extensive research progress in VAE-based self-supervised multi-view clustering and multi-modal learning methods \cite{1,6,77,113,114,116,117,118,119,120,136}. A pioneering self-supervised generative clustering method within the VAE framework was introduced by Jiang et al. \cite{115}.
However, existing methods often grapple with challenges related to large-scale datasets and suboptimal sample reconstruction. In response to these issues, Yin et al. presented DMVCVAE \cite{119}, a novel multi-view clustering method that learns a shared generative latent representation conforming to a mixture of Gaussian distributions. In a similar vein, Xu et al. proposed Multi-VAE \cite{1}, capable of acquiring disentangled and interpretable visual representations, thereby addressing large-scale data clustering problems. 
Different from the existing multi-view clustering methods, they introduce a view-common variable $y$ and multiple view-peculiar variables $\left\{\mathbf{z}^{(1)}, \mathbf{z}^{(2)}, \ldots, \mathbf{z}^{(V)}\right\}$ in a multiple VAE architecture. The model can disentangle all views’ common cluster representations and each view’s peculiar visual representations. In this way, the interference of multiple views’ superfluous information is reduced when mining their complementary information for clustering. The generative model of Multi-VAE can be expressed as follows:
\begin{equation}
\begin{aligned}
p\left(\mathbf{x}^{(v)}, \mathbf{z}^{(v)}, \mathbf{y}\right) & =p\left(\mathbf{x}^{(v)} \mid \mathbf{z}^{(v)}, \mathbf{y}\right) p\left(\mathbf{z}^{(v)}, \mathbf{y}\right) \\
& =p\left(\mathbf{x}^{(v)} \mid \mathbf{z}^{(v)}, \mathbf{y}\right) p\left(\mathbf{z}^{(v)}\right) p(\mathbf{y}) ,
\end{aligned}
\end{equation}
the $\mathbf{x}^{(v)}$ denotes the data for all views and $\mathbf{c}$ denotes the view public variable. For the $v$-th view, $\mathbf{z}^{(v)}$ denotes its unique visual information. The posteriors of $\mathbf{y}$ and $\mathbf{z}^{(v)}$ are written as $p\left(\mathbf{y} \mid\mathbf{x}^{(v)}\right)$ and $p\left(\mathbf{z}^{(v)} \mid \mathbf{x}^{(v)}\right)$. 

VAE has not only made significant strides in the realm of self-supervised MVC but has also yielded impressive results in multi-modal representation learning. DMMVAE \cite{117}, for instance, introduces a generative variational model capable of learning both the private and shared latent spaces for each modality. Each latent variable corresponds to a disentangled representational factor. The model enhances inter-modal compatibility by introducing a cross-VAE task, aiming for cross-modal reconstruction through a shared latent space.
In summary, VAE facilitates the mapping of data into a latent space through collaborative learning involving an encoder and a generator. It can generate high-quality data samples from the latent distribution. VAE's primary focus lies in capturing the structure of data distribution, maximizing data likelihood, and minimizing the KL divergence of the latent distribution during training, all of which contribute to feature learning and data generation. Differing from GAN, VAE places more emphasis on the continuity of data distribution and its feature learning capabilities, which confer advantages in fields such as multi-view clustering and multi-modal learning.

\subsection{Contrastive Methods}
\begin{figure}[h]
\centering
\includegraphics[width=1\linewidth]{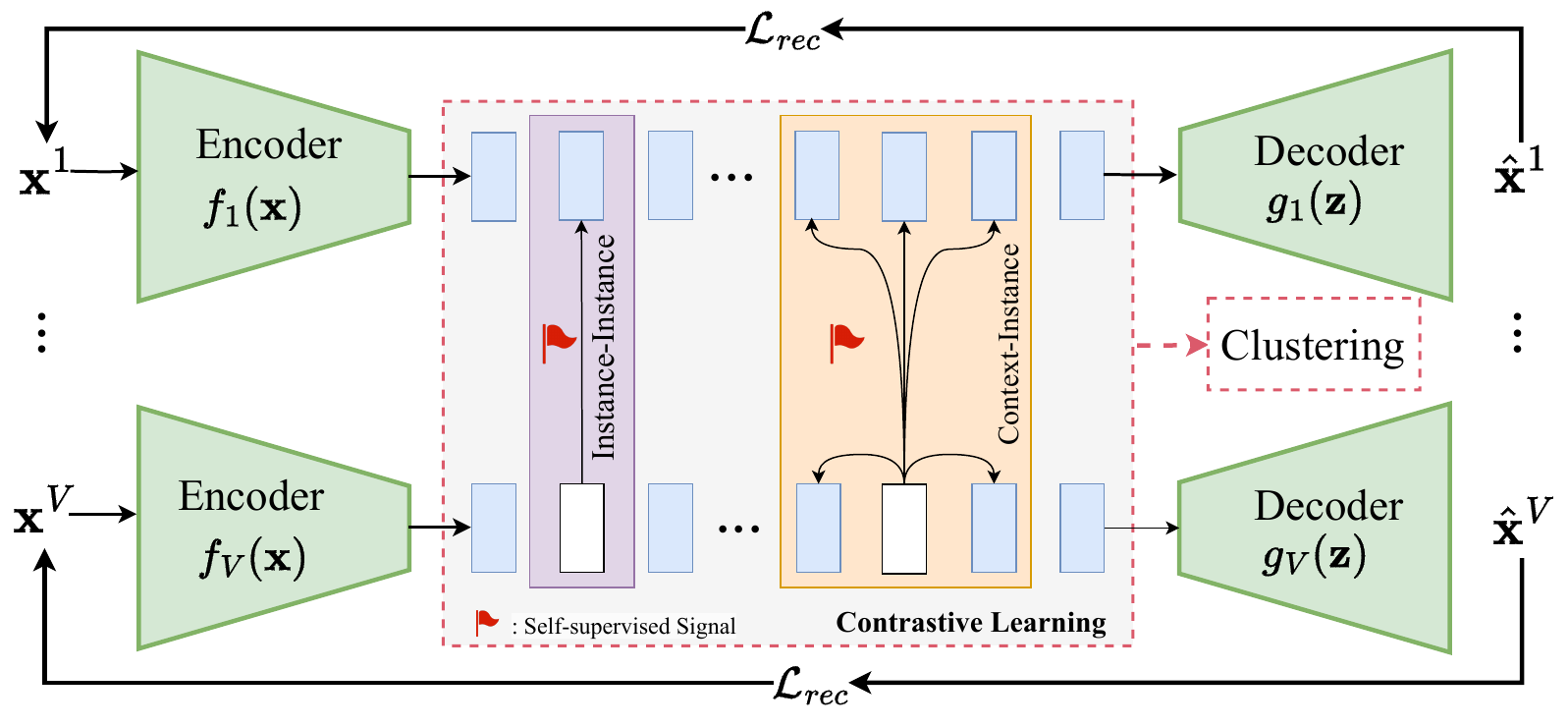}
\caption{Each encoder-decoder denotes the processing of a view, and the red dashed box denotes the contrastive learning module, where the purple portion denotes the instance-instance and the orange portion denotes the context-instance. Red flags denote the self-supervised signals required for the clustering process.}
\label{fig12}
\end{figure}

To address the challenges posed by heterogeneity, noise, and dimensional inconsistencies in multi-view clustering, the integration of contrastive learning into multi-view clustering is recognized as an effective method. Its primary objective is to enhance clustering performance by bolstering the consistency among different views. Contrastive learning, as a paradigm, endeavors to acquire meaningful feature representations of data by assessing distinctions between samples, maximizing similarities among similar samples across different views, and minimizing similarities between dissimilar samples. This paradigm finds widespread use in self-supervised learning, exemplified by techniques like CMC \cite{148}, MoCo \cite{147}, and SimCLR \cite{149}.
As illustrated in Fig. \ref{fig12}, we can categorize contrastive methods into two main groups: Instance-Instance and Context-Instance.

\subsubsection{Instance-Instance}

Instance-instance contrastive learning centers on the features of individual samples and assesses the similarities and differences among samples to improve feature distinctiveness. Given the inherent data representation inconsistencies among different views in multi-view data, namely, heterogeneity, instance-instance contrastive learning becomes particularly valuable in capturing correlation information between samples. This method reinforces cross-view consistency by bringing similar samples closer in feature representations through pairs of samples from different views. We can define the formula for instance-instance contrastive learning adoption as follows:

\begin{equation}
\begin{aligned}
\mathcal{L}_{\text {Instance-Instance }}=-\log \frac{\exp \left(\operatorname{sim}\left(\mathbf{z}_{i}, \mathbf{z}_{j}\right) / \tau\right)}{\sum_{k=1}^{N} 1_{[k \neq i]} \exp \left(\operatorname{sim}\left(\mathbf{z}_{i}, \mathbf{z}_{k}\right) / \tau\right)},
\end{aligned}
\end{equation}
where $\mathbf{z}_{i}$ and $\mathbf{z}_{j}$ are the feature representations of the sample instances after feature extraction, $sim(·)$ denotes the similarity measure between $\mathbf{z}_{i}$ and $\mathbf{z}_{j}$, $N$ is the total number of samples,  $\tau$ is the temperature parameter to control data distribution, and $l_{[k \neq i]}$ is the indicator function, which takes the value of 1 when $k \neq i$ and 0 otherwise.

%
Numerous studies have underscored the significance of instance-instance contrastive learning in the context of multi-view clustering \cite{5,67,68,141,37,142,19,7,17}, with InstDisc \cite{151} serving as a prominent example. Building upon InstDisc, CMC \cite{148} introduces a novel method by considering multiple different views of an image as positive samples while designating another view as the negative counterpart. CMC seeks to bring together multiple views of an image in the embedding space while distancing them from other samples. Specifically, only one negative sample is selected for each positive sample. MoCo \cite{147} further advances this idea by introducing the concept of momentum comparison, significantly increasing the number of negative samples. However, MoCo's method of positive sampling is somewhat simplistic; each pair of positive samples originates from the same source without any transformation or augmentation, making these positive samples easily distinguishable. In contrast, SimCLR \cite{149} underscores the importance of utilizing a challenging positive sampling strategy by introducing up to 10 forms of data augmentation. This augmentation method shares similarities with CMC, which leverages various views to enhance positive sample pairs. On a different note, BYOL \cite{150} adopts a more radical method by forgoing negative sampling altogether in self-supervised learning, achieving superior results.

The advancement of the aforementioned contrastive learning techniques has spurred extensive research efforts within the domain of multi-view clustering.
Fang et al. introduced an inductive multi-view image clustering framework incorporating self-supervised contrastive heterogeneous graph co-learning \cite{16}. This framework incorporates two contrastive objectives, aiming to merge multiple views, achieve comprehensive local feature propagation embedding, and maximize mutual information between local feature propagation and influence-perception feature propagation. Nonetheless, the method falls short of fully harnessing the latent complementary information inherent in different views.
In a related vein, SCMC \cite{26}, as proposed by Zhang et al., leverages view-specific autoencoders to transform raw multi-view data into compact features representing perceptually nonlinear structures. Recognizing the challenges posed by missing views and view inconsistency in real-world scenarios, Wang et al. integrated multi-view information to align data and acquire latent representations, unveiling a novel cross-view graph comparison learning framework \cite{145}.
To address noise issues encountered in previous comparison methods, Yang et al. identified noise within those methods and introduced the use of known correspondences to construct positive pairs and random sampling for constructing negative pairs \cite{124}. However, despite previous studies addressing multiple targets within the same feature space, they often overlooked the conflict between learning consistent public semantics and reconstructing inconsistent views of private information.
Subsequently, Xu et al. introduced a new framework for multi-level feature learning in contrastive multi-view clustering \cite{2}. This framework learns various levels of features, encompassing low-level features, high-level features, and semantic labels/features, from the original features without fusion, effectively achieving both the reconstruction and consistency goals across different feature spaces. Moreover, in the high-level feature space, the framework strengthens the consistency objective through contrastive learning, enabling high-level features to concentrate on learning the common semantics shared among all views.

\subsubsection{Context-Instance}

Context-instance contrastive learning places emphasis on understanding the attribution relationship between a sample's local features and its global context \cite{124,126,140,144,145,2,16,26}. In multi-view applications, different views offer unique contextual information, leading to richer feature representations when instances are compared across these views. More specifically, positive sample pairs are formed by pairing local features or global context from the same sample, while negative sample pairs are generated using different global contexts. The ultimate objective is to maximize similarity between positive samples in the feature space while minimizing disparities between positive and negative samples. We define the generalized formula for Context-Instance comparison learning as follows:
\begin{equation}
\begin{aligned}
\max _{f} \sum_{i=1}^{N} \log \frac{e^{\operatorname{sim}\left(f\left(\mathbf{x}_{i}\right), f\left(\mathbf{x}_{j}\right)\right)}}{\sum_{k=1}^{K} e^{\operatorname{sim}\left(f\left(\mathbf{x}_{i}\right), f\left(\mathbf{x}_{k}\right)\right)}},
\end{aligned}
\end{equation}
where $\mathbf{x}_i$ denotes the $i$-th sample, $\mathbf{x}_j$ denotes a positive sample associated with sample $\mathbf{x}_i$, $\mathbf{x}_k$ denotes a negative sample that is different from $\mathbf{x}_i$, $f(·)$ denotes the representation of the sample in the feature space, $N$ denotes the total number of samples, $K$ denotes the total number of negative samples, and $sim(·)$ denotes the similarity measure between the samples.

Many earlier studies, such as ACTIVE \cite{67} proposed by Wang et al., sought to mitigate the impact of individual missing data on clustering. Meanwhile, Kaveh Hassani et al. \cite{19} focused on learning node and layer representations by comparing structural views of graphs. In contrast to these methods, Pan et al. introduced contrastive loss as a regularization technique in MCGC \cite{68} to enhance the feasibility of consensus graph clustering. Specifically, contrastive learning is utilized to maximize mutual information among different views, resulting in information-rich and consistent representations. Dual prediction is employed to minimize the conditional entropy of different views, facilitating the recovery of missing views.
Furthermore, Ke et al. \cite{141} introduced a novel contrastive fusion technique, which explores multi-view alignment from the perspective of information bottlenecks and aligns the specific representation of each view through the introduction of intermediate variables. Lin et al. \cite{5} amalgamated representation learning and data recovery into a unified framework from an information-theoretic standpoint. Differing from the majority of existing contrastive learning research, these studies directly aim to maximize mutual information between different view representations. These contributions play a vital role in consolidating the theoretical framework of Context-Instance contrastive learning recovery. They highlight researchers' endeavors to explore diverse methods to address the clustering challenges posed by multi-view data, resulting in significant advancements in this domain.

\subsection{Generative-Contrastive Methods}
Generative-contrast learning, also known as adversarial learning, combines the features of both generative and contrastive learning and has gained significant traction in the field of multi-view clustering \cite{146, 5, 121}. On one hand, in adversarial learning, generative learning still preserves the underlying generator structure, which includes encoders and decoders. It focuses on reconstructing the original data distributions rather than individual samples by minimizing differences. This addresses the inherent limitations of pointwise generative learning, particularly when dealing with noisy or high-dimensional data. It also imparts unique expressiveness to adversarial learning generative models.
On the other hand, compared to contrastive learning, which relies solely on distinguishable information to differentiate between various samples, adversarial learning can yield more effective feature representations. This is because it encompasses all the essential information required to construct the inputs. We define the generalized formula for generative-contrast learning as follows:

\begin{equation}
\begin{aligned}
\min _{G} \max _{D} \mathbb{E}_{\mathbf{x} \sim P_{\text {data }}}[\log D(\mathbf{x})]+\mathbb{E}_{\mathbf{z} \sim P_{\text {noise }}}[\log (1-D(G(\mathbf{z})))],
\end{aligned}
\label{18}
\end{equation}
where $P_{\text {data }}$ denotes the distribution of real data, $P_{\text {noise }}$ denotes the input of noise or other generators, $\mathbb{E}_{\mathbf{x} \sim P_{\text {data }}}\log D(\mathbf{x})]$ denotes the error of the discriminator D on the real sample x, $\mathbb{E}_{\mathbf{z} \sim P_{\text {noise }}}[\log (1-D(G(\mathbf{z})))]$ denotes the error of the discriminator $D$ on the generated sample $G(\mathbf{z})$, $G(\mathbf{z}) $denotes the sample generated by the generator $G$ through the input of noise or other generators, $D(\mathbf{x})$ denotes the probability that the discriminator $D$ will classify the input sample $\mathbf{x}$ as true, and $D(G(\mathbf{z}))$ denotes the probability that the discriminator $D$ will classify the generated sample $G(\mathbf{z})$ as true.

Lin et al. \cite{5} have integrated representation learning and data recovery into a unified framework from an information-theoretic perspective. In this method, information-rich and consistent representations are obtained by maximizing mutual information between different views through contrastive learning. Simultaneously, missing views are recovered by minimizing the conditional entropy of different views through pairwise prediction.
In contrast, the CIMIC-GAN \cite{146}, proposed by Jiatai Wang et al., employs a generative adversarial network to populate incomplete data and employs double contrastive learning to learn the consistency between complete and incomplete data. This method specifically incorporates the automatic encoding representation of both complete and incomplete data into double-contrastive learning to achieve consistency in learning. It takes into account the diversity and complementarity of multiple modalities. By incorporating GAN into the auto-encoding process, the model not only effectively leverages the novel features of incomplete data but also enhances its generalization in scenarios with a high rate of missing data

\subsection{Others}

Self-supervised MVC extends beyond the traditional paradigms of generative, contrastive, or generative-contrastive learning. There are methods that blend different concepts or employ alternative strategies to fully exploit the inherent relevance of multi-view data. These methods strive to achieve more accurate and robust clustering results, forging new paths in the field of multi-view clustering and continually advancing its progress and development.
Some multi-view clustering methods depart from traditional generative or contrastive learning techniques. Instead, they harness alternative self-supervised signals to facilitate consistency learning and clustering. These methods employ strategies such as data augmentation, similarity propagation, or self-agreement mechanisms to bolster the alignment of representations across different views, ultimately leading to enhanced clustering outcomes. For example, in \cite{4}, the proposal involves constructing in-depth feature learning and clustering models for each view independently, effectively leveraging the complementary information within multi-view data. CMVC \cite{143} takes a different method by normalizing the views to enhance the features learned from each view, ultimately improving clustering quality through optimized training strategies.

\section{Future Work And Discussion}\label{sec5}
In this section, we delve into the advantages of self-supervised multi-view clustering while also addressing several open issues and suggesting potential directions for future research.

In the realm of real-world data, diversity abounds, with various modalities and views presenting themselves. Self-supervised multi-view clustering endeavors to enhance clustering performance by amalgamating information from distinct data modalities. This method offers several notable advantages:
\begin{itemize}
    \item Self-supervised multi-view clustering excels at capturing richer data features and uncovering correlations among different views by integrating information from multiple data sources, thereby enhancing the quality of clustering results.

    \item In real-world scenarios with numerous missing data, self-supervised multi-view clustering exhibits the ability to mitigate the impact of data incompleteness to a certain extent. Whether through inferring missing information or making judgments based on available data, self-supervised multi-view clustering can still yield meaningful clustering results in situations of data scarcity.
    
    \item Self-supervised multi-view clustering is adept at exploiting the consistency and alignment present among multi-view data, leading to more precise identification and analysis of underlying patterns and structures within the data. This, in turn, enables a deeper understanding and provides valuable insights.
\end{itemize}

However, despite the remarkable achievements of self-supervised multi-view clustering in enhancing clustering performance, several challenges persist, leaving room for potential improvements:
\begin{itemize}
    \item Current research frequently relies on experiments conducted with artificial incomplete datasets, which to some extent, limits its applicability and reliability in real-world scenarios. Conducting more empirical studies using real-world data can provide better validation of the method's effectiveness.

    \item Self-supervised multi-view clustering still grapples with the challenge of accurately handling incomplete data. Existing methods may exhibit sensitivity to missing data situations, necessitating the development of more robust missing data recovery and filling strategies to bolster clustering resilience.
    
    \item Inherent data biases and noise-related challenges continue to impact the quality of clustering results, especially in the presence of noisy data. Future research endeavors can explore methods for more effectively addressing these issues to achieve more accurate and robust clustering outcomes.
\end{itemize}

In summary, self-supervised multi-view clustering holds significant promise for a wide range of applications in multi-modal data analysis. However, several technical and practical challenges must be addressed to further enhance its performance and applicability.

\section{Conclusion}\label{sec6}
The self-supervised learning problem presents a significant challenge within the realm of MVC, and its investigation holds paramount importance for practical applications. To provide readers with a comprehensive grasp of self-supervised MVC, we acquaint them with primary research materials within related fields. This includes commonly employed self-supervised MVC datasets and related problems, offering insights from both image and video perspectives.
Subsequently, this paper introduces a novel classification method aimed at categorizing existing self-supervised MVC methods: representation learning and self-supervised learning. 
Representation learning plays an integral role within self-supervised MVC and the process of generating and utilizing self-supervised signals. We place our focus on the identification and categorization of self-supervised signals into five distinct categories based on specific methodologies, providing detailed descriptions for each.
Self-supervised learning methods encompass two distinct learning models that leverage the input data itself as supervision. These models are generation-based and comparison-based. Generative methods strive to learn the underlying data distribution and represent the data using generative models. In contrast, comparison methods directly optimize an objective function that involves pairwise similarity, aiming to minimize the average similarity within clusters while maximizing the average similarity between clusters.
Finally, it is imperative to highlight several open and challenging problems, encouraging researchers to delve deeper into further research and make substantial progress in this domain.
\section*{Acknowledgments}
This work  was supported by the National Science Foundation of China (61962045, 62062055, 61902382, 61972381), Program for Young Talents of Science and Technology in Universities of Inner Mongolia Autonomous Region(NJYT23104), the Science and Technology Planning Project of Inner Mongolia Autonomous Region (2023YFSH0066).
\medskip


\end{document}